\documentclass[letterpaper, 10 pt, conference]{ieeeconf}
\IEEEoverridecommandlockouts 
\usepackage{amsmath}
\usepackage{amsfonts}
\usepackage{graphicx}
\usepackage{hyperref}
\usepackage{cite}
\usepackage{float}
\usepackage[caption=false,font=footnotesize]{subfig}
\usepackage{xcolor}
\usepackage[b]{esvect}
\usepackage{fancyhdr}
\usepackage{etoolbox}
\usepackage{mdframed}
\usepackage{array}
\usepackage{xcolor}

\newcommand{\col}[1]{\underline{#1}}

\renewcommand{\bullet}{\boldsymbol{\cdot}}

\graphicspath{{img/},{fig/}}

\begin{document}

\title{\LARGE \bf
Model-Based Real-Time Pose and Sag Estimation of Overhead Power Lines Using LiDAR for Drone Inspection
}

\author{ Alexandre Girard$^{1}$, Steven A. Parkison$^{2}$, Philippe Hamelin$^{2}$ 
\thanks{$^{1}$Alexandre Girard is with the Department of Mechanical Engineering, Universite de Sherbrooke, Qc, Canada {\tt\small alex.girard@usherbrooke.ca }}
\thanks{$^{2}$Steven A. Parkison and Philippe Hamelin are with the Hydro-Québec Research Institute (IREQ), Qc, Canada. }
}%

\maketitle
\thispagestyle{empty}
\pagestyle{empty}

\begin{abstract}
Drones can inspect overhead power lines while they remain energized, significantly simplifying the inspection process. However, localizing a drone relative to all conductors using an onboard LiDAR sensor presents several challenges: (1) conductors provide minimal surface for LiDAR beams limiting the number of conductor points in a scan, (2) not all conductors are consistently detected, and (3) distinguishing LiDAR points corresponding to conductors from other objects, such as trees and pylons, is difficult. This paper proposes an estimation approach that minimizes the error between LiDAR measurements and a single geometric model representing the entire conductor array, rather than tracking individual conductors separately. Experimental results, using data from a power line drone inspection, demonstrate that this method achieves accurate tracking, with a solver converging under 50 ms per frame, even in the presence of partial observations, noise, and outliers. A sensitivity analysis shows that the estimation approach can tolerate up to twice as many outlier points as valid conductors measurements. 
\end{abstract}

\section{Introduction}





Population growth, the emergence of new electricity-powered technologies, and even climate change are all factors contributing to the constant increase in energy demand. This compels electric utilities to optimize the operation of their network, from production to transmission and distribution. To offer a level of reliability that meets customer expectations, the inspection and maintenance of these critical infrastructures become important issues. Among these, transmission power lines are particularly critical, as they are both the backbone of the network and subject to environmental hazards.

\begin{figure}[t]
\begin{center}
\includegraphics[width=0.99\linewidth]{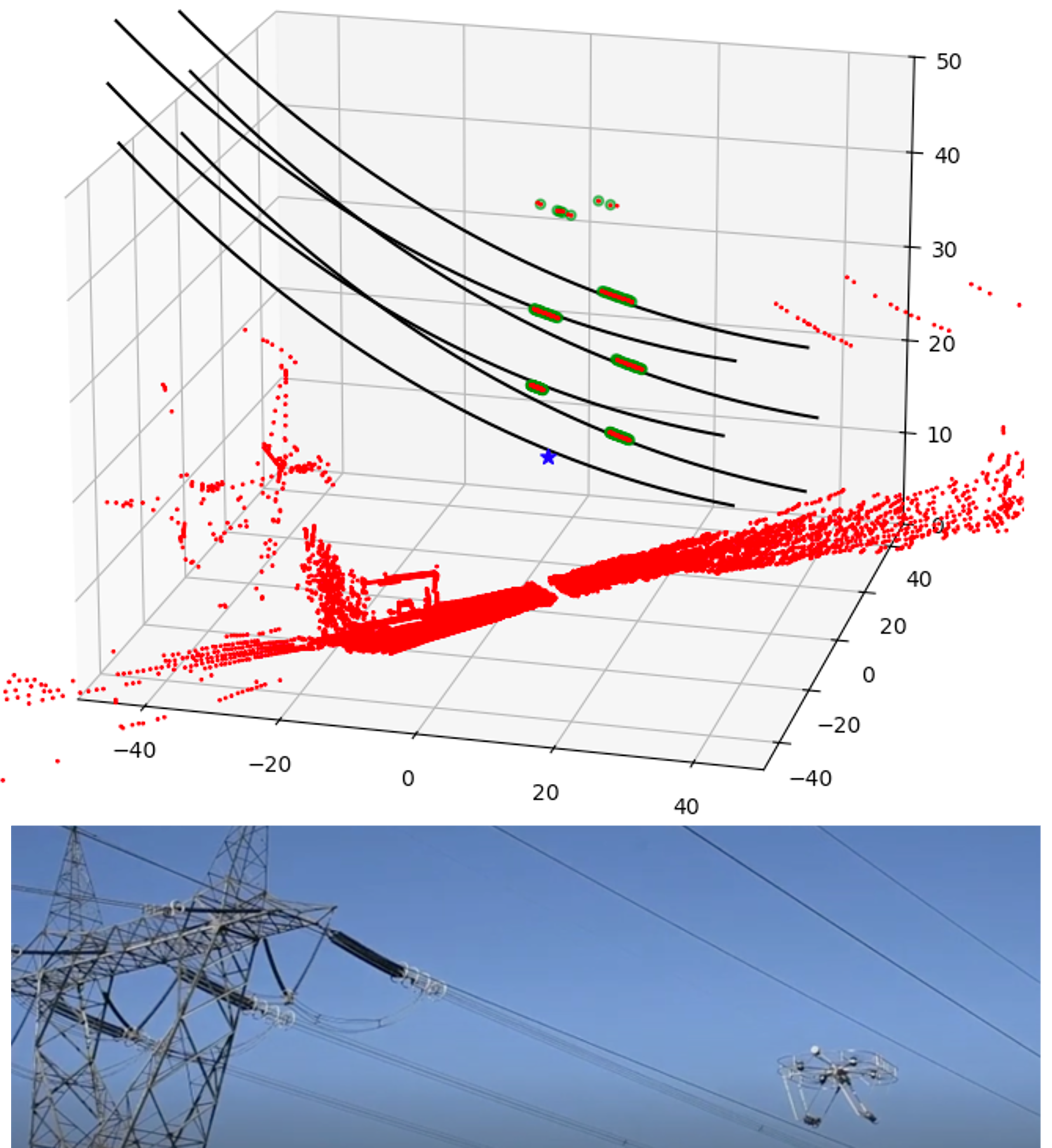}
\vspace{-20pt}
\caption{A scene from an experimental test of a drone inspecting a 315kV power line. The figure above shows one frame with the raw LiDAR measurements (red dots), filtered points (green dots), the estimated power line catenary curves (black), and the drone position (blue star); all axis are meters.}
\label{fig:exp_global}
\vspace{-10pt}
\end{center}
\end{figure}

Recently, drones have revolutionized the field of transmission power line inspection and maintenance, allowing tasks to be performed more efficiently and safely. This field is largely dominated by visual inspection applications, where the drone remains at a safe distance from the line \cite{burde2024}. However, a new niche has emerged in recent years: hybrid drones capable of landing and rolling on live power lines \cite{miralles2018}, opening the door to new applications such as contact inspection \cite{pouliot2020}, deployment of probes on conductors \cite{hamelin2024}, and maintenance. This capability introduces the complex requirement of accurately identifying, localizing, and navigating in close proximity to individual conductors. This paper details an estimation scheme to determine power line pose and intrinsic geometric parameters during these advanced inspection missions, such as a contact inspection with the LineDrone \cite{linedrone} illustrated at Fig. \ref{fig:exp_global}.

\subsection{Background}

%
%
%
%
%


Power line detection approaches can be categorized based on several criteria: the sensing modality used, whether they operate offline or in real-time, the techniques employed for classifying conductors, and the methods used to parameterize the shape and position of the conductors.


The two main sensing modalities used to inspect power lines are cameras and light detection and ranging (LiDAR) sensors. Cameras have proven useful when combined with fast, real-time UAV control algorithms for power line detection \cite{zhou2016robust,xing2023perception_aware_mpc}. Zhang et al. demonstrated that recent advances in convolutional neural networks (CNNs) can be used to detect power lines in images \cite{zhang2019convolution}. Madaan et al. \cite{madaan_multi-view_2019} presented an approach to reconstruct power lines using images taken from multiple angles. However, cameras lack the rich 3D information available from LiDAR sensors, which can be used to detect power lines \cite{mclaughlin,azevedo,Sohn,Paneque,Liang,Zhang,Tan,Wang,Yermo}.


Aerial surveying with LiDAR sensors has been shown to be a useful tool for offline power line detection and monitoring, as surveyed by Shen et al. \cite{SHEN2024104056}. Examples of offline approaches include McLaughlin \cite{mclaughlin}, who merged point clouds from a helicopter-mounted LiDAR using GNSS data, and Sohn et al. \cite{Sohn}, who also used data from a helicopter-mounted LiDAR that made one pass over a transmission line corridor. Real-time detection of power lines can be desirable \cite{azevedo,Paneque,SCHOFIELD2022pl_detction_and_tracking}, to provide feedback to a human operator or control algorithms \cite{zhou2016robust,xing2023perception_aware_mpc}.


A variety of approaches have been investigated to detect and classify power lines. A popular strategy involves clustering points to distinguish features such as ground, pylons, and conductors \cite{mclaughlin,azevedo,Sohn,Liang,Zhang,Tan, Wang, Yermo, SCHOFIELD2022pl_detction_and_tracking}. However, these methods often assume a global observation of the conductors and rely on offline analysis, which can be time-consuming. The Hough line transform has also been investigated to classify conductors \cite{zhou2016robust} \cite{Paneque}, but it requires reducing the problem to a 2D classification task. Modern neural network techniques from the AI community have also been explored \cite{xing2023perception_aware_mpc, zhang2019convolution} \cite{madaan_multi-view_2019}, but these supervised approaches require labeled training data that match the modality of the data being used, and therefore often necessitate camera data.


The final criterion that differentiates existing literature on power line detection is the geometric parametrization of the conductors themselves. Linear models can be used \cite{SCHOFIELD2022pl_detction_and_tracking}, especially when combined with camera data \cite{zhou2016robust, xing2023perception_aware_mpc}. However, linear approaches lose critical information about the sag in the conductors, which is important for applying global navigation algorithms and extrapolating the shape of the conductor in partial observation scenarios. Polynomial parametrization can also be employed to better represent the conductor \cite{Liang2011_polynomial}, but it is still not perfectly accurate, as a hanging catenary is actually a hyperbolic cosine. This more accurate representation has been used by others to parameterize detected conductors \cite{madaan_multi-view_2019, mclaughlin, azevedo, Sohn}.

\subsection{Novel contribution}

In this paper, we introduce a novel algorithm for detecting conductors from airborne LiDAR data, applicable to both global (when measurement points from the whole span of a power line are available) and partial (when measurement points are only available from a short local section) observations. Our approach leverages a multi-conductor geometric model based on the hyperbolic cosine catenary equation, enabling the classification of conductors relative to an a priori configuration, and also to always estimate the position of all conductors even when one of of them is not observed by the LiDAR. This capability is crucial for autonomous drone navigation, particularly to enable the drone to navigate to the conductor specifically identified by the mission. The proposed algorithm executes in less than 50 ms, thereby opening the door to real-time onboard processing on a drone. 




\section{Estimation algorithm}

\subsection{Geometric model}

The proposed approach for the estimation problem is to leverage geometrical and physical assumptions to greatly reduce the number of parameters required to fully determine the space occupied by an array of conductors. Under reasonable physical assumptions for a typical overhead power line, the shape of suspended conductors will follow the catenary equation given by
\begin{equation}
z_j = cat( x_j | a ) = a \left[ \cosh{ \left( \frac{x_j}{a} \right)  } - 1 \right]
\label{eq:cate}
\end{equation}
that describe the relationship between a vertical position $z_j$ and an horizontal position $x_j$ based on a parameter $a$ defining the sag, when expressed in a local frame whose origin is the lowest point on the catenary curve (that could be a virtual point not physically on a conductor in some situation like steep ascent/descent).

\begin{figure}[tpb]
\begin{center}
\includegraphics[width=0.99\linewidth]{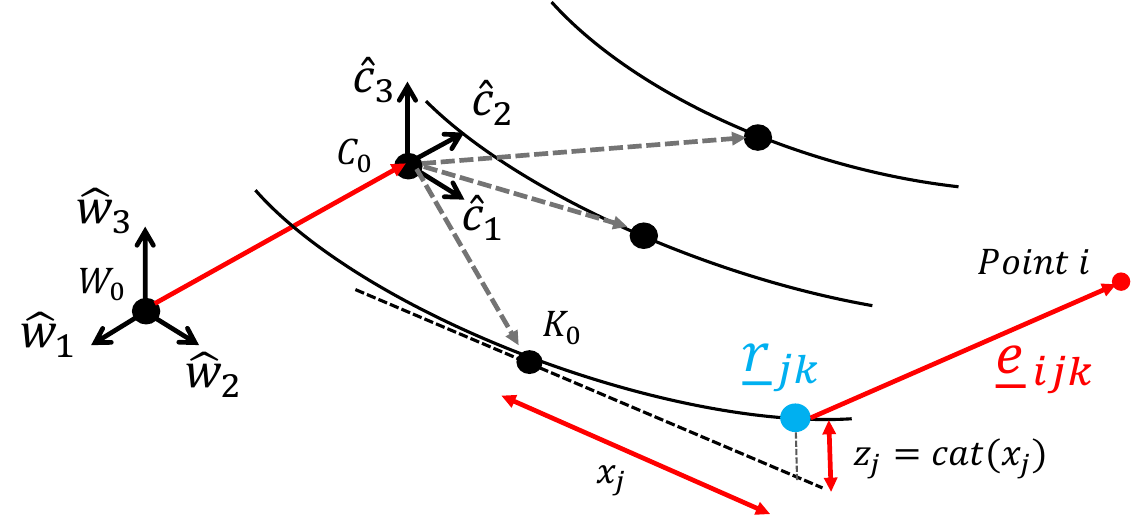}
\vspace{-10pt}
\caption{Geometric model for a group of conductors, and error vector of a measurement point $i$ with respect to a model point $jk$ (position $x_j$ on conductor $k$).}\label{fig:array}
\vspace{-10pt}
\end{center}
\end{figure}

\begin{figure}[h]
\vspace{-20pt}
\begin{center}
\includegraphics[width=0.75\linewidth]{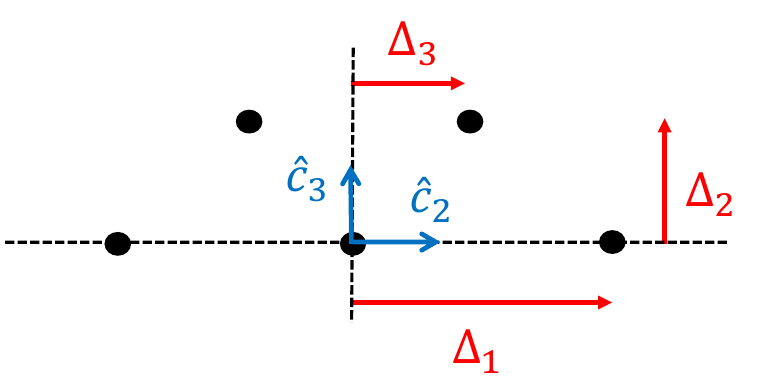}
\vspace{-10pt}
\caption{Parametrization of the internal translations in a conductor array}\label{fig:32}
\vspace{-10pt}
\end{center}
\end{figure}

Overhead power lines are organized in groups of conductors sharing an orientation and sag parameter, up to usually relatively small variations. The proposed model, illustrated in Fig. \ref{fig:array}, splits the parametrization of the pose of an intermediary power line frame $\{C_o,\hat{c}_1,\hat{c}_2,\hat{c}_3\}$ with respect to an arbitrary world frame $\{W_o,\hat{w}_1,\hat{w}_2,\hat{w}_3\}$, and local translations of each line catenary origin $K_j$ with respect to $C_o$. This parametrization is useful for defining search bounds since it is possible to use prior knowledge of the power line configuration to constrain distances between conductors. The following geometric equation describe the position of points occupied by conductors in a world frame:
\begin{align}
\col{r}_{jk/W_o}^{w} &= {}^wR^c \, \left(  \col{r}_{j/K_o}^{c} + \col{r}_{K_o/C_o}^{c} \right) + \col{r}_{C_o/W_o}^{w}
\\
\begin{bmatrix}
    x_{jk}^w \\
    y_{jk}^w \\
    z_{jk}^w
\end{bmatrix} &= 
\begin{bmatrix}
    c\psi & -s\psi & 0 \\
    s\psi &  c\psi & 0 \\
    0         &  0         & 1 
\end{bmatrix}
\,
\begin{bmatrix}
x_j + x_k\\ y_k \\ cat( x_j , a ) + z_k
\end{bmatrix}
+
\begin{bmatrix}
x_o \\ y_o \\ z_o 
\end{bmatrix}
\label{eq:arraymodel}
\end{align}
where $k$ is an index for the conductor in the array, and $j$ is an index associated with a specific point given a local horizontal position $x_j$ relative to the catenary origin. This equation has five global parameters ($x_o$, $y_o$ $z_o$, $\psi$ and $a$) corresponding to 3 translations, one rotation about $\hat{w}_3$ and the sag parameter $a$, plus three internal translations ($x_k$, $y_k$ $z_k$) per conductor in the array. 
A matrix $M$ encoding all the translation vector $\col{r}_{K_o/C_o}^{c}$ as a function of a limited number of $l$ internal offset parameters is introduced:
\begin{equation}
\begin{bmatrix}
x_1 & \hdots & x_k  & \hdots & x_q \\ 
y_1 & \hdots & y_k  & \hdots & y_q \\ 
z_1 & \hdots & z_k  & \hdots & z_q \\ 
\end{bmatrix}
= M( 
\underbrace{
\Delta_1 \; \hdots \; \Delta_l
}_{ \text{internal offsets} }
)
\label{eq:offsetsfunc}
\end{equation}
where $q$ is the number of conductors in the array. Given the known configuration of a power line and its symmetries, a reduced set of offset $\Delta$ parameters is sufficient to determine this matrix. For example, a standard configuration of conductors, illustrated at Fig. \ref{fig:32}, is usually symmetric and can be parametrize with only 3 $\Delta$ parameters:
\begin{equation}
M_{32}(\Delta_1,\Delta_2,\Delta_3) = 
\scriptsize
\begin{bmatrix}
0 & 0 & 0 & 0 & 0   \\ 
-\Delta_1 & 0 & \Delta_1 & -\Delta_3 & \Delta_3   \\
0 & 0 & 0 & \Delta_2 & \Delta_2 
\label{eq:matrixmodel}
\end{bmatrix}
\end{equation}

Integrating eq. \eqref{eq:offsetsfunc} into eq. \eqref{eq:arraymodel}, we have now a forward geometric model for computing world points:
\begin{equation}
\col{r}_{jk/W_o}^{w} = f( x_j | \; \col{p} \; ) 
\end{equation}
as a function of a vector of $(5+l)$ parameters:
\begin{equation}
\col{p}^T = \left[ 
\underbrace{
x_o \; y_o \; z_o \; \psi
}_{ \text{pose} } \;
\underbrace{
a
}_{ \text{sag} } \;  
\underbrace{
\Delta_1 \; \hdots \; \Delta_l
}_{ \text{internal offsets} }
\right]
\end{equation}
and a free variable $x_j$ describing a position along the powerline in the $\hat{c}_1$ direction.

\subsection{Loss function}

The estimation problem is formulated as searching for the vector of parameters $\col{p}$ that minimizes a loss function, which weighs the costs associated with measurement points and includes a regularization term that penalizes changes in the parameter vector relative to a previous estimate $\hat{\col{p}}$:
\begin{align}
J( \col{p} ) = 
\underbrace{ 
R^T \col{c} 
}_{ \text{points cost} }
+ 
\underbrace{ 
(\hat{\col{p}} - \col{p})^T \; Q \; (\hat{\col{p}} - \col{p})
}_{ \text{ regulation term } }
\end{align}
where $\col{c}$ is a vector where each element $c_i$ is a cost associated with the measurement $i$, $R$ is a vector of weights and $Q$ is a matrix of weights (parametrizing the tradeoff between trusting measurements vs. the model like in a Kalman filter). 

The cost $c_i$ of a point, is based on the distance $d_i$ to the nearest conductor that can be computed using the geometric model. However, a logarithm cost function is used:
\begin{align}
c_i = \log \left( 1 + ( d_i )^{2} \right)
\end{align}
This is sometimes referred to as a Lorentzian cost function, which behaves similarly to a quadratic function for small distances but tapers off for larger distances, reducing the influence of outliers \cite{fitzgibbon_robust_2003}.

\begin{figure}[htpb]
\vspace{-10pt}
\begin{center}
\includegraphics[width=0.6\linewidth]{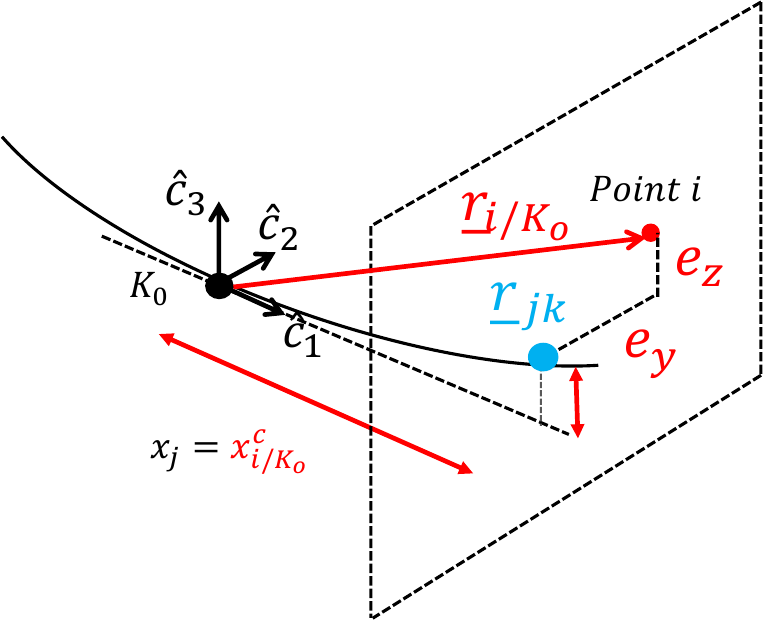}
\caption{The method of associating a model point $jk$ (position $x_j$ on the conductor $k$) with a measurement point $i$.}\label{fig:x_method}
\end{center}
\end{figure}

In order to compute the distance $d_i$, as illustrated in Fig. \ref{fig:array}, for each measurement point $i$, an error vector $\col{e}_{ijk}$ is defined for each conductor $k$ and a coordinate $x_j$:
\begin{align}
\col{e}_{ijk} &= 
\underbrace{ 
\col{r}_{i/W_o}
}_{ \text{Measurement point $i$ } }
- 
\underbrace{ 
\col{r}_{jk/W_o}( x_j | \; \col{p} \; ) 
}_{ \text{Model point $j$ on line $k$ } }
\end{align}
The error vector can be expressed in the catenary frame basis:
\begin{align}
\col{e}_{ijk}^c &= 
\scriptsize
\begin{bmatrix}
    c\psi &  s\psi & 0 \\
   -s\psi &  c\psi & 0 \\
    0     &  0     & 1 
\end{bmatrix}
\begin{bmatrix}
x_i - x_o \\ y_i - y_o \\ z_i - z_o
\end{bmatrix}
-
\begin{bmatrix}
x_k \\ y_k \\ z_k
\end{bmatrix}
-
\begin{bmatrix}
x_j \\ 0 \\ cat( x_j | a ) 
\end{bmatrix}
\end{align}
The distance to the nearest conductor $d_i$ of a measurement point $\col{r}_{i}^{w}$ in the world frame is the minimum error over all conductors (index $k$) and position along the conductor (index $j$):
\begin{align}
d_i( \col{p} ) &= 
\operatornamewithlimits{min}\limits_{k} 
\left[ 
\operatornamewithlimits{min}\limits_{j} 
\| \col{e}_{ijk}^w \|
\right] \\
d_i( \col{p} ) &= 
\operatornamewithlimits{min}\limits_{k} 
\underbrace{ \left[ 
\operatornamewithlimits{min}\limits_{j} 
\|\col{r}_{i}^{w} -  \col{r}_{jk}^{w}( x_j | \; \col{p} ) \|
\right]}_{\text{Closet model point on line $k$}}
\label{eq:distancesamplemethod}
\end{align}
While computing the minimum over a small number of conductors is fast, sampling and computing over all positions $x_j$ along the powerline is computationally demanding. To associate the closest point along a line, a simplifying assumption leveraging the fact that conductors are at small angle with respect to the direction $\hat{c}_1$ is used. As illustrated in Fig. \ref{fig:x_method}, the scheme is to select the model point $j$ corresponding to the intersection between the model curve and the plane formed by 1) the measured point $i$ and 2) a normal direction defined by $\hat{c}_1$. This also correspond to using the $x$ coordinate of the measurement (with respect to the vertex point $k_o$ in the catenary vector basis) as the $x_j$ variable.

It is thus possible to compute $x_j$ as a function of measurement and model parameters:
\begin{align}
x_{j} &= 
c\psi ( x_i - x_o ) +
s\psi ( y_i - y_o ) - x_k
\end{align}
and to substitute it in the error equation, leading to:
\begin{align}
\col{e}_{ik}^c &= 
\begin{bmatrix}
0 \\ 
-s\psi ( x_i - x_o ) +
 c\psi ( y_i - y_o ) - y_k
\\ 
z_i - z_o - z_k - z_j
\end{bmatrix}
\label{eq:errorx}
\end{align}
with 
\begin{align}
z_j &= 
a \left[ \cosh{ \left( \frac{
c\psi ( x_i - x_o ) + s\psi ( y_i - y_o ) - x_k
}{a} \right)  } - 1 \right]
\end{align}
With this scheme, computing the distance of a measurement to the model, only a minimum over the number of conductors in the group is required:
\begin{align}
d_i( \col{p} ) &= 
\operatornamewithlimits{min}\limits_{k} 
\left[  
\| \col{e}_{ik}^c \|
\right] 
\label{eq:distancesamplemethod2}
\end{align}
Note that the associated conductor $k$ to a measurement point will be fixed to avoid the discontinuity when computing the gradient of this cost function. 

\subsection{Mathematical program and solver}

The estimation of geometric parameters best fitting a cloud point is formulated as the following constrained optimization:
\begin{equation}
\begin{aligned}
\col{\hat{p}}_{new} = 
&\operatornamewithlimits{min}\limits_{\col{p}} 
J( \, \col{p} \, , \, PTS \, , \, \col{\hat{p}} \, , \, R \, , \, Q \, ) 
\\
&
\textrm{s.t.} \quad p_i^{l} \leq p_i \leq p_i^{u}
\end{aligned}
\end{equation}
where $PTS$ is a point cloud of $m$ 3D measurements:
\begin{equation}
PTS = 
\begin{bmatrix}
x_1 &  & x_i & & x_m \\  y_1  & \hdots & y_i & \hdots & y_m \\ z_1 & & 
z_i 
& & z_m
\end{bmatrix}
\end{equation}
$\col{\hat{p}}$ is a prior estimate, $\col{p}^{l}$ and $\col{p}^{u}$ are vectors of lower and upper bound on the geometric parameters.

A sequential quadratic programming algorithm (SQP) is then used to solve this mathematical program, but with an additional search step to help avoid local minima. A total of $n_{search}+1$ SQP programs are launched at each steps with different initial guess. One SQP program is using the previous estimate to warm start the solver, while the others use the previous estimate plus a random perturbation following a normal distribution $~\mathcal{N}(0,\,\Sigma)$, where $\Sigma$ is a diagonal covariance matrix. The solution with the lowest cost of all SQP program is kept. All in all, the full estimation algorithm has the following hyperparameters: 
\begin{equation}
    \theta = \{
R , Q , \col{p}^{l}, \col{p}^{u}, n_{search}, \Sigma \}
\end{equation}

An open source implementation of this algorithm in Python, using Scipy SQP solver \cite{noauthor_scipyoptimizeminimize_nodate}, integrating a wide range of power line configurations and providing the gradients of cost functions to the solver was developed and is available at the following link: \url{https://github.com/alx87grd/catenary}.

\section{Performance analysis}

Two types of analysis are presented: a sensibility analysis with simulated LiDAR data studying the robustness to outlier points, and an experimental validation using data from an inspection on an energized 315kV power line.

\subsection{Simulated data}
\begin{figure}[htbp]
\vspace{-10pt}
\begin{center}
\includegraphics[trim={0 0.5cm 0 0.25cm},clip,width=0.99\linewidth]{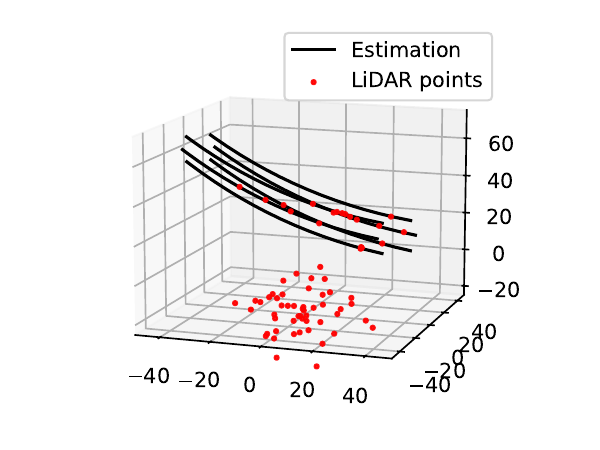}
\vspace{-15pt}
\caption{Sample of points for the global observations task with $n_{o}=50$ outlier points. In the black the resulting estimated catenary curves after 100 frames. Axes represent distances in meters.}
\vspace{-10pt}
\label{fig:global_obs}
\end{center}
\end{figure}

\begin{figure}[htbp]
\vspace{-10pt}
\begin{center}
\includegraphics[trim={0 0.5cm 0 0.25cm},clip,width=0.99\linewidth]{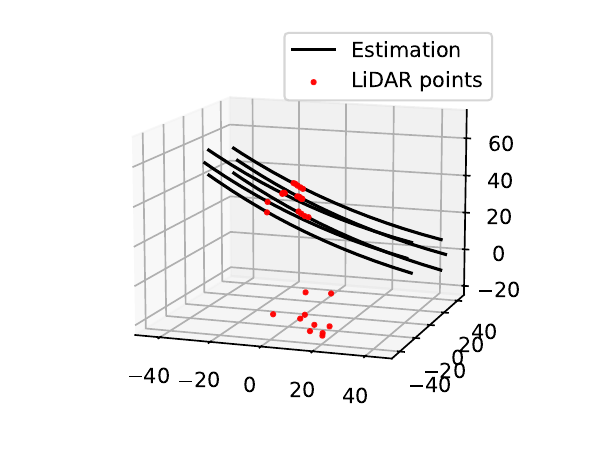}
\vspace{-15pt}
\caption{Sample of points for the partial observations task with $n_{o}=10$ outlier points. In the black the resulting estimated catenary curves after 100 frames. Axes represent distances in meters.}
\label{fig:partial_obs}
\vspace{-10pt}
\end{center}
\end{figure}

Two simulated scenarios are evaluated: 1) when measurements are available from the whole length of the catenary curve, see Fig. \ref{fig:global_obs} (analogous to when the drone LiDAR scan plane is mostly parallel to the conductors, as it can happen during the approach phase, or if accumulating an history of points in a buffer) and 2) when measurements are only available momentary from a local slice, see Fig. \ref{fig:partial_obs} (analogous to when the drone LiDAR scan plane is perpendicular to the conductors, as is it the case during the insertion phase of our inspection scenario). The synthetic LiDAR measurements are created by sampling a number (uniform distribution between 0 and 10) of points per line using the true equation with an additive noise (normal distribution), plus additional $n_{o}$ outliers points generated from an independent cluster (analogous to the data from real mission in the presence of high trees or pylons). The evaluation is done for 100 frame of data, and repeated 100 times with random initial parameters and seeds to get statistical performance results. The code to reproduce the simulated results and all the hyper-parameters used, are available at this link \url{https://github.com/alx87grd/catenary}.

\begin{figure}[htbp]
\vspace{-10pt}
\begin{center}
\includegraphics[trim={0 0.25cm 0 0},clip,width=0.99\linewidth]{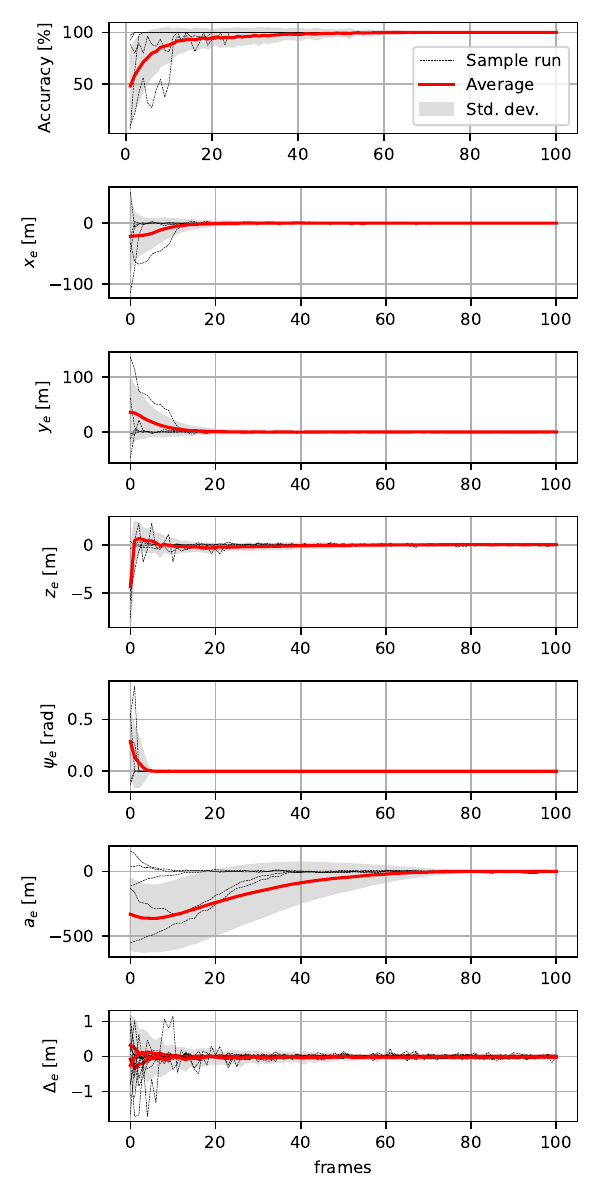}
\vspace{-10pt}
\caption{Estimation performance (global observations and 10 outliers).}
\vspace{-10pt}
\label{fig:n_in_sim_global}
\end{center}
\end{figure}

\begin{figure}[htbp]
\vspace{-10pt}
\begin{center}
\includegraphics[trim={0 0.25cm 0 0},clip,width=0.99\linewidth]{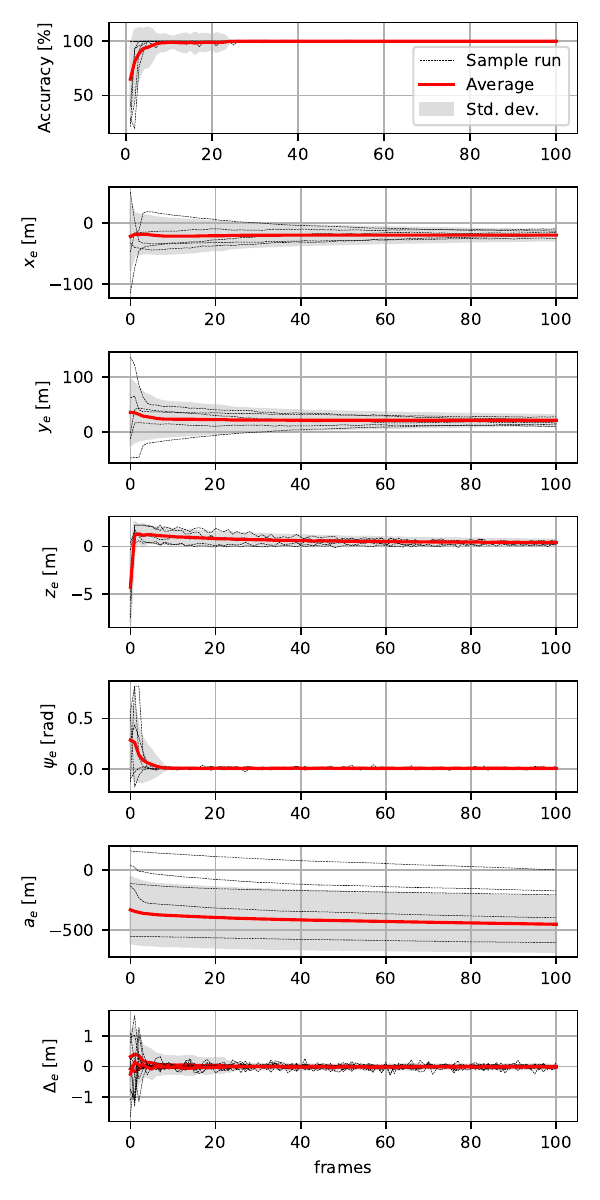}
\vspace{-10pt}
\caption{Estimation performance (partial observations and 10 outliers).}
\label{fig:n_in_sim_partial}
\end{center}
\end{figure}

Fig. \ref{fig:global_obs} and Fig. \ref{fig:partial_obs} show the estimated catenary curve of the conductors after 100 frame for two sample scenarios. Fig. \ref{fig:n_in_sim_global} and Fig. \ref{fig:n_in_sim_partial} show statistical results for the 100 runs, when the number of outliers is set to $n_{o}=10$. The accuracy metric is defined as follow: it is the ratio of the number of points in the data that are explained by the model using the estimated parameters (under one meter of the projected conductors curves) divided by the number of points explained by the model using the ground truth parameters. In other words, an accuracy of 100\% means that the estimation model fit as many points as the true projected curves. This definition is used to avoid having to label each points in the experimental data test and keeping the same accuracy metric for both simulated and experimental analysis. Fig. \ref{fig:n_in_sim_global} and Fig. \ref{fig:n_in_sim_partial} also show the evolution of the error on the estimated parameters, note that all the error on internal offset parameters $\Delta_1$, $\Delta_2$ and $\Delta_3$ are grouped on the same subplot for brevity. For partial observations, while all real measurement are explained after 100 frame, not all parameters converge to the ground truth values to accomplish this. Indeed, there is a whole subset of parameters leading to catenaries that pass roughly through the points. 

\subsection{Sensitivity to outliers points}

A sensitivity analysis was conducted to evaluate the impact of the number of outlier points ($n_{o}$) on the estimation algorithm's performance. The last 10 LiDAR frames of the simulation were used to compute performance metrics. Tables \ref{tab:glo} and \ref{tab:par} summarize the tracking performance for global and partial observations, respectively. In these tables, $n_{o}$ represents the number of outlier points, $n_{pts}$ is the average number of points sent to the estimation algorithm in each frame, $\Delta t$ is the average solver convergence time per SQP program (computed on a personal computer with a Apple M4 Max chip), Acc. is the average accuracy as described earlier, $\psi$ is the average error on the orientation parameter, and $a$ is the average error on the sag parameter. In both cases, the accuracy of the estimation starts to drop when the number of outlier points exceeds 75, which is more than twice the average number of real points on the conductors, approximately 30 on average in the tests.

\begin{table}[th]
   \centering 
   \caption{Tracking performance (global) vs. outliers} 
   \label{tab:glo}
   \begin{tabular}{c c c c c c}
   \hline
    \noalign{\smallskip} \noalign{\smallskip} \noalign{\smallskip}
   $n_{o}$ & $n_{pts}$ & $\Delta t$ [ms] & Acc. [\%] & $\psi_e$ [] & $a_e$ [m] \\ \noalign{\smallskip} \noalign{\smallskip} \hline \noalign{\smallskip}  \noalign{\smallskip}  
    10  &  40 +/- 6 &  6 +/- 1  & 100 +/- 0 & 0 +/- 0 &   -0 +/- 6   \\
      20  &  50 +/- 6 &  6 +/- 1  & 100 +/- 0 & 0 +/- 0 &  -2 +/- 16   \\
      50  &  80 +/- 6 &  7 +/- 1  &  98 +/- 6 &   0 +/- 0    & -79 +/- 160  \\
      75  & 105 +/- 6 &  7 +/- 1  & 94 +/- 11 &   0 +/- 0    & -174 +/- 254 \\
     100  & 130 +/- 6 &  8 +/- 1  & 87 +/- 17 &   0 +/- 0    & -261 +/- 299 \\
     150  & 180 +/- 6 &  9 +/- 1  & 68 +/- 22 &   0 +/- 0    & -401 +/- 278 \\
     200  & 230 +/- 6 &  10 +/- 2 & 54 +/- 16 &   0 +/- 0    & -470 +/- 214 \\
     300  & 330 +/- 6 &  10 +/- 1 & 50 +/- 16 &   0 +/- 0    & -484 +/- 174 \\
   \hline
   \end{tabular}
\end{table}
\begin{table}[th]
   \centering 
   \caption{Tracking performance (partial) vs. outliers} 
   \label{tab:par}
   \begin{tabular}{c c c c c c}
   \hline
    \noalign{\smallskip} \noalign{\smallskip} \noalign{\smallskip}
   $n_{o}$ & $n_{pts}$ & $\Delta t$ [ms] & Acc. [\%] & $\psi_e$ [] & $a_e$ [m] \\ \noalign{\smallskip} \noalign{\smallskip} \hline \noalign{\smallskip}  \noalign{\smallskip} 
      10  &  40 +/- 6 &  7 +/- 1  & 100 +/- 0 & 0 +/- 0  & -449 +/- 233 \\
      20  &  50 +/- 6 &  7 +/- 1  & 100 +/- 1 & 0 +/- 0  & -444 +/- 223 \\
      50  &  80 +/- 6 &  7 +/- 1  & 97 +/- 11 & 0 +/- 0  & -395 +/- 231 \\
      75  & 105 +/- 6 &  7 +/- 1  & 96 +/- 12 & 01 +/- 0 & -374 +/- 239 \\
     100  & 130 +/- 6 &  8 +/- 1  & 93 +/- 17 & 01 +/- 0 & -344 +/- 250 \\
     150  & 180 +/- 6 &  8 +/- 1  & 90 +/- 17 & 01 +/- 0 & -328 +/- 262 \\
     200  & 230 +/- 6 &  9 +/- 2  & 86 +/- 19 & 01 +/- 0 & -323 +/- 266 \\
     300  & 330 +/- 6 &  10 +/- 1 &  83 +/- 8 & 01 +/- 0 & -335 +/- 267 \\
   \hline
   \end{tabular}
\end{table}

\subsection{Experimental tests with an inspection drone}

The estimation algorithm was evaluated using an experimental dataset collected during a mission in which the drone took off from the ground and approached the energized power line. The drone was a modified LineDrone \cite{linedrone} equipped with a Velodyne Puck Lite 3D LiDAR and a NVIDIA Jetson Xavier NX computer. For this mission, the LiDAR scan plane was primarily perpendicular to the power line, resulting in very partial observations of the conductors. The mission targeted a double-circuit power line, composed of six conductors and two ground wires. Here for the test, the geometric model was defined to estimate only the six conductors, and LiDAR returns from the two ground wires are thus noisy outliers to filter out. To evaluate the performance of the estimation algorithm, ground truth parameters of the model were extracted by visual inspection of the LiDAR point cloud.

Although the proposed algorithm is robust to some extent against outliers, it is necessary to apply a filter to extract the points belonging to the conductors, as the experimental point cloud contains points from the ground, pylons, and other objects in the environment. To this end, three different filters were applied to the experimental dataset:

\begin{enumerate}
    \item \emph{Corridor}: This filter leverages prior information about the approximate position of the pylons to include only points within a rectangular area between the pylons. An elevation histogram is used to remove ground points.
    \item \emph{Ground}: This filter removes ground points using the RANSAC \cite{ransac} plane fitting algorithm. Simpler ground plane estimation methods were found to be perform poorly mostly due to returns from the vegetation.
    \item \emph{Clustering}: This filter involves the following steps: a) ground points are removed using the aforementioned ground filter, b) the DBSCAN algorithm \cite{dbscan} is used to extract clusters, and c) only clusters with a shape similar to a line are retained by examining the eigenvalues of the point cloud covariance matrix. 
\end{enumerate}

\begin{figure}[htb]
\vspace{-10pt}
\begin{center}
\includegraphics[width=0.99\linewidth]{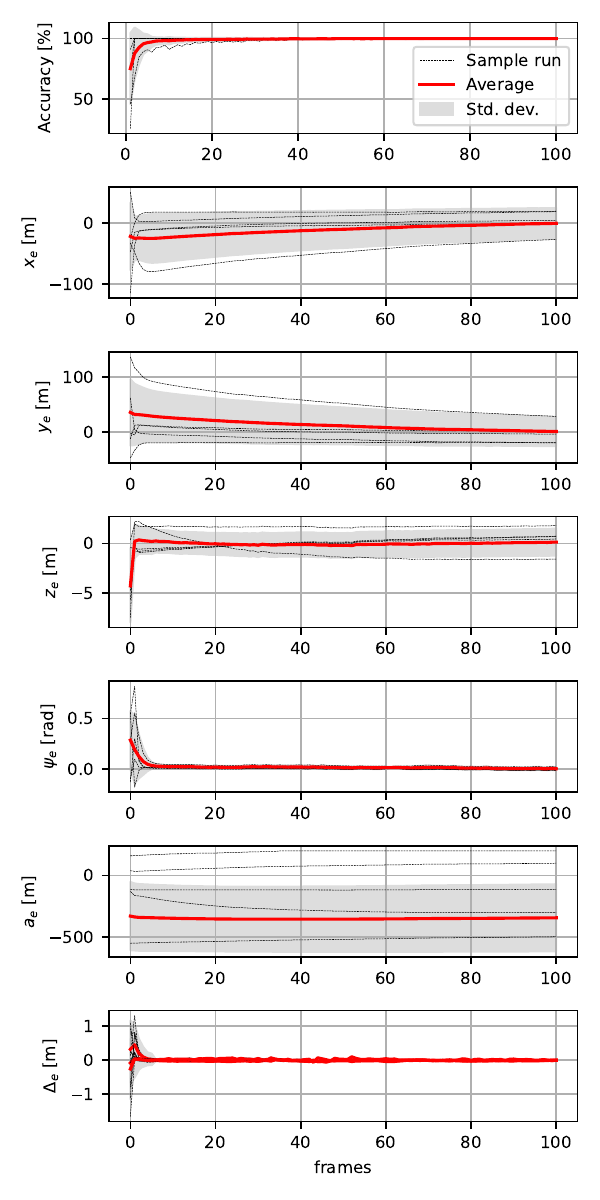}
\vspace{-10pt}
\caption{Performance with the experimental data with the corridor filter.}
\label{fig:exp_a}
\end{center}
\end{figure}

Fig. \ref{fig:exp_global} illustrates the application of the \emph{Corridor} filter for a single LiDAR frame. In this figure, the red points represent the LiDAR data before the filtering process, while the green points represent the data after filtering. The green points are subsequently used by the estimation algorithm. The results of the estimation algorithm for this filter are detailed in Figure \ref{fig:exp_a}. It is shown that within 20 frames, the algorithm converges to parameters that fully explain the LiDAR points. The orientation $\psi$ and internal offsets parameters are found to converge to the ground truth. However, the translation and sag parameters do not converge to the ground truth as they are not fully constrained by the partial observations. Nevertheless, it still allows the drone to have a sufficient local approximation of the power line configuration for navigation purposes.


\begin{figure}[htb]
\begin{center}
\includegraphics[trim={0 0.5cm 0 0.25cm},clip,width=0.99\linewidth]{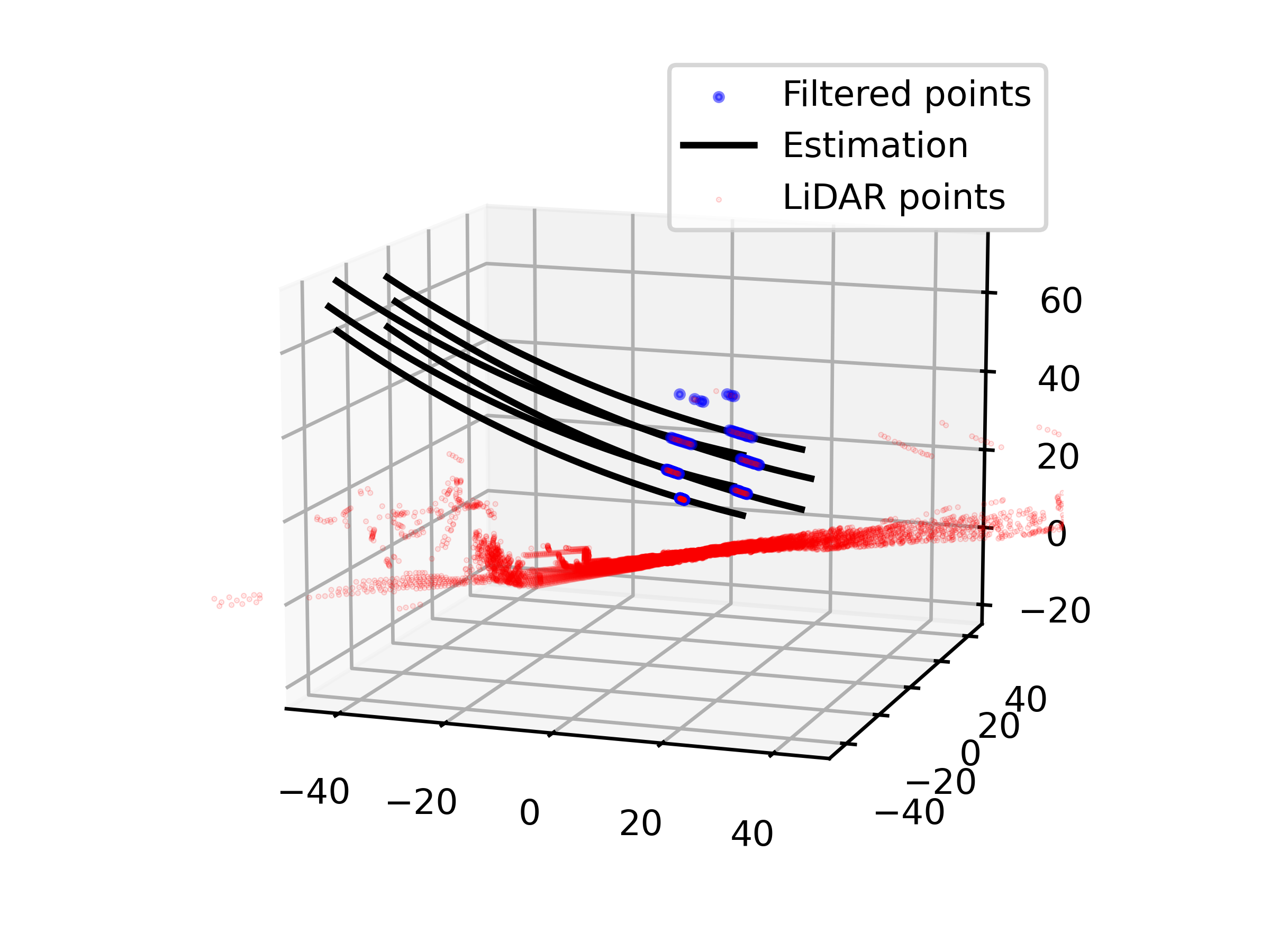}
\vspace{-10pt}
\caption{Estimation with the corridor filter, axes are distances in meters.}
\label{fig:corridor}
\vspace{-10pt}
\end{center}
\end{figure}
\begin{figure}[htb]
\begin{center}
\includegraphics[trim={0 0.5cm 0 0.25cm},clip,width=0.99\linewidth]{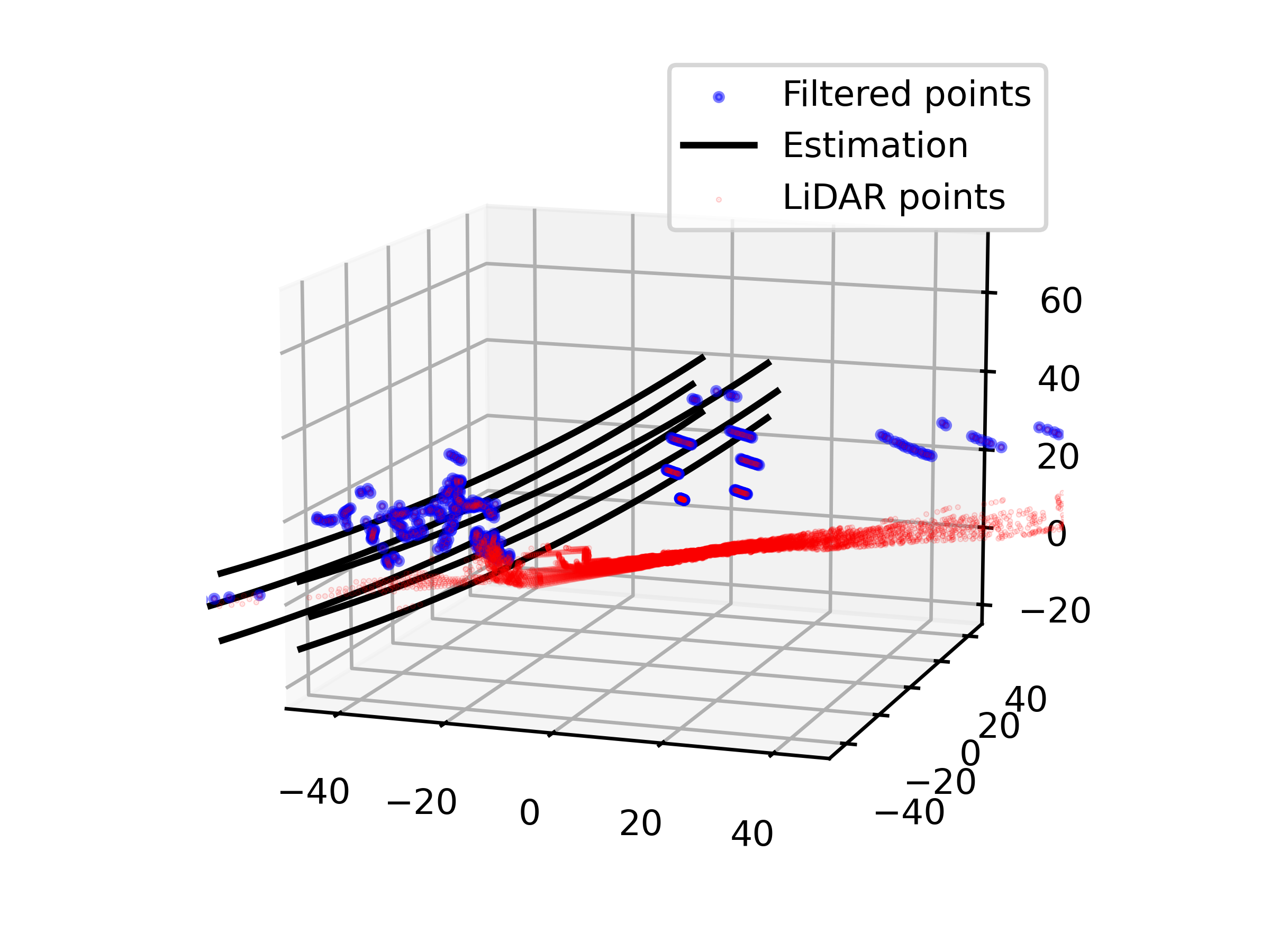}
\vspace{-10pt}
\caption{Estimation with the ground filter, axes are distances in meters.}
\label{fig:ground}
\vspace{-10pt}
\end{center}
\end{figure}
\begin{figure}[htb]
\vspace{-10pt}
\begin{center}
\includegraphics[trim={0 0.5cm 0 0.25cm},clip,width=0.99\linewidth]{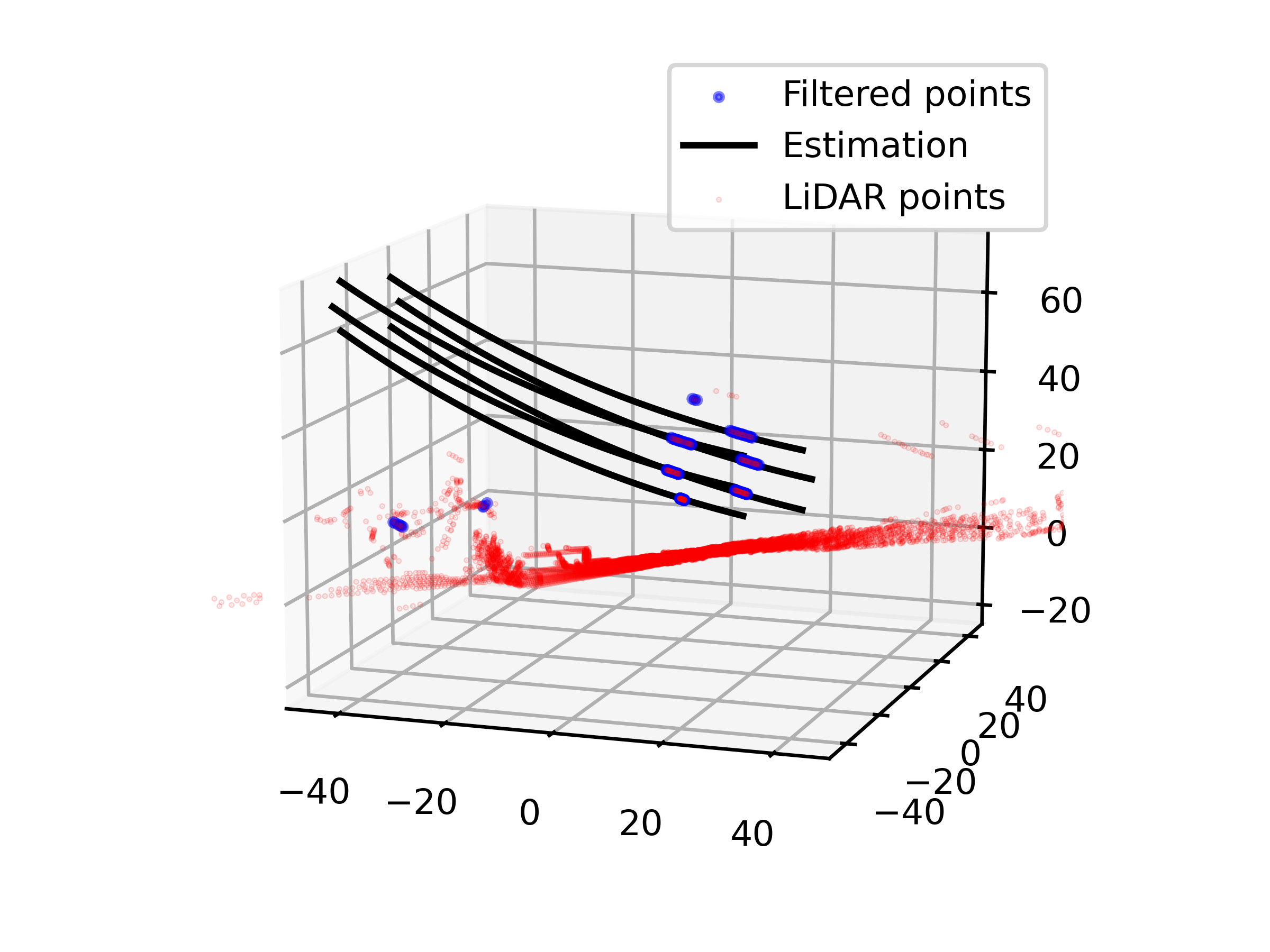}
\vspace{-10pt}
\caption{Estimation with the clustering filter, axes are distances in meters.}
\vspace{-10pt}
\label{fig:clustering}
\end{center}
\end{figure}

The experimental dataset was also used to compare the performance of the proposed algorithm when using three different filters. Figures \ref{fig:corridor}, \ref{fig:ground}, and \ref{fig:clustering} illustrate the final estimated catenaries for the \emph{Corridor}, \emph{Ground}, and \emph{Clustering} filters, respectively. Additionally, Table \ref{tab:exp} summarizes the statistical results for all three filters. The results show that both the \emph{Corridor} and \emph{Clustering} filters effectively remove enough outliers for the estimation algorithm to properly converge on the solution, whereas the \emph{Ground} filter leads to the failure of the estimation. In this dataset, a large number of LiDAR points originated from a nearby pylon, and the \emph{Ground} filter was unable to remove these points. Consequently, the estimated solution converged toward the pylon, as the number of points from the pylon was significantly larger than the number of points from the conductors, approximately 500 points versus 100 points, highlighting a limit of the algorithm in a practical scenario, i.e. tackling inspections when multiple power lines are relatively close. 

\begin{table}[tph]
   \centering 
   \caption{Tracking performance vs. filter type} 
   \label{tab:exp}
   \begin{tabular}{c c c c } 
   \hline
    \noalign{\smallskip} \noalign{\smallskip} \noalign{\smallskip}
   Filter & $n_{pts}$ & $\Delta t$ [ms] & Acc. [\%] 
   \\ \noalign{\smallskip} \noalign{\smallskip} \hline \noalign{\smallskip}  \noalign{\smallskip} 
        Corridor   & 172 +/- 7  &  9 +/- 1  & 100 +/- 0 \\
 Ground & 609 +/- 47 &  9 +/- 1  &  12 +/- 3 \\
   Clustering  & 185 +/- 14 &  9 +/- 1  & 100 +/- 0 \\
    \hline
   \end{tabular}
   \vspace{-10pt}
\end{table}

\subsection{Discussion}
Additional tests and simple ablation studies were also conducted to reach a better understanding of the algorithm performance and limitations.
\textbf{Outliers distribution:} The tolerance to outliers was found to be much more robust if they are uniformly distributed in the space, like in rain or snow, instead of clustered in small zones, like an unfiltered pylon. 
\textbf{Hyper-parameters:} It was found that the algorithm works well for a wide range of hyper-parameters, however two schemes are essential: \textbf{1)} the search steps with a perturbed initial guess is found to be necessary to get out of local minima created by the multiple conductors and group of points, \textbf{2)}the logarithm relationship between error and cost is essential for the algorithm to work in the presence of outliers. 
\textbf{Data associated simplification:} Using sampling and brute force for associating the closest model point was tested and found to have the same performance (but slower) compared to the simplification presented at eq. \eqref{eq:errorx}.


\section{Conclusion}

In conclusion, this paper presents a novel algorithm for real-time estimation of the pose, offsets, and sag of power lines using a non-linear programming approach. The proposed method leverages a multi-conductor geometric model based on the hyperbolic cosine catenary equation, enabling accurate estimation of conductors in both global and partial observation scenarios. The algorithm demonstrates robust performance, with a solver convergence time of approximately 30-50 ms per frame (on a personal computer processor), even in the presence of noise and outlier points. Sensitivity analysis confirms the algorithm's ability to tolerate up to twice the number of outlier points as measurement points corresponding to the conductors. Experimental validation using data from a drone inspecting a power line further showcases the effectiveness of the approach for power line inspection and maintenance missions. The method also allows to keep tack of all conductors despite occlusions.  The main limitation of the algorithm is that it require prior knowledge of the configuration of the power line, which can be done manually for an inspection mission.  
However in future work it would be interesting to combine the tracking ability presented in this paper with an automatic identification of the configuration for a more general usage.

\bibliographystyle{IEEEtran}
\bibliography{catenary}

\newpage

\appendix

\begin{table}[ht]
   \centering 
   \caption{Variables} 
   \label{expVari}
   \begin{tabular}{c p{4.2cm} }
   \hline
   \hline \noalign{\smallskip} \noalign{\smallskip} \noalign{\smallskip}
   \textbf{Variables} & \textbf{Descriptions} \\ \noalign{\smallskip} \noalign{\smallskip} \hline \hline \noalign{\smallskip} 
   \multicolumn{2}{c}{\textbf{Model parameters (to be estimated)}} \\ \noalign{\smallskip}  \hline \hline \noalign{\smallskip} 
   $x_o$ & x translation of the catenary frame in the world frame \\ \noalign{\smallskip} \hline \noalign{\smallskip}
   $y_o$ & x translation of the catenary frame in the world frame \\ \noalign{\smallskip} \hline \noalign{\smallskip}
   $z_o$ & x translation of the catenary frame in the world frame \\ \noalign{\smallskip} \hline \noalign{\smallskip}
   $\psi$ & z rotation of the catenary frame relative to the world frame \\ \noalign{\smallskip} \hline \noalign{\smallskip}
   $a$ & sag parameter of the catenary curve \\ \noalign{\smallskip} \hline \noalign{\smallskip}
   $\Delta_1$, ... , $\Delta_l$ & internal offset parameters (for an array model) \\ \noalign{\smallskip} \hline \noalign{\smallskip}
   \multicolumn{2}{c}{\textbf{Measurement}} \\ \noalign{\smallskip}  \hline \hline \noalign{\smallskip} 
   $\col{r}_{i/W_o}^w= \begin{bmatrix} x_i \\ y_i \\ z_i \end{bmatrix}$ & 3D point $i$ measured with respect to the origin $W_o$ in vector basis $w$ \\ \noalign{\smallskip} \hline \noalign{\smallskip}
   \multicolumn{2}{c}{\textbf{Model kinematic}} \\ \noalign{\smallskip}  \hline \hline \noalign{\smallskip} 
   $ w = \{ \hat{w}_1 , \hat{w}_2 , \hat{w}_3 \}$ & world frame vector basis \\ \noalign{\smallskip} \hline \noalign{\smallskip}
   $ c = \{ \hat{c}_1 , \hat{c}_2 , \hat{c}_3 \}$ & catenary frame vector basis \\ \noalign{\smallskip} \hline \noalign{\smallskip}
   $ W_o $ & world frame origin\\ \noalign{\smallskip} \hline \noalign{\smallskip}
   $ C_o $ & local catenary frame origin\\ \noalign{\smallskip} \hline \noalign{\smallskip}
   $\col{r}_{j/K_o}^c = \begin{bmatrix} x_j \\ 0 \\ z_j \end{bmatrix}$ & position of points $j$ on the catenary curve relative to its vertex point $K_0$ expressed in $c$ vector basis \\ \noalign{\smallskip} \hline \noalign{\smallskip}
   $\col{r}_{C_o/W_o}^w = \begin{bmatrix} x_o \\ y_o \\ z_o \end{bmatrix}$& translation of the catenary frame in the world frame  \\ \noalign{\smallskip} \hline \noalign{\smallskip}
   $\col{r}_{K_o/C_o}^c = \begin{bmatrix} x_k \\ y_k \\ z_k \end{bmatrix}$& translation of the vertex of conductor $k$ in local catenary frame   \\ 
    \noalign{\smallskip} \hline \hline \noalign{\smallskip}
    ${}^wR^c = \begin{bmatrix}
    \cos \psi & -\sin \psi & 0 \\
    \sin \psi &  \cos \psi & 0 \\
    0         &  0         & 1 
    \end{bmatrix}$ 
    & rotation matrix of catenary vector basis with respect to the world vector basis \\
    \noalign{\smallskip} \hline \hline \noalign{\smallskip}
    \multicolumn{2}{c}{\textbf{Cost function}} \\ \noalign{\smallskip}  \hline \hline \noalign{\smallskip} 
   $\col{e}_{ijk} = \col{r}_{i/jk} $ & error between measurement $i$ and model point $j$ on conductor $k$ \\ \noalign{\smallskip} \hline \noalign{\smallskip}
   $d_{i} = \sqrt{ \col{e}_{i{j^*}}^T\col{e}_{i{j^*}} } $ & distance between measurement $i$ and associated model point $j^*$ \\ 
    \noalign{\smallskip} \hline \hline \noalign{\smallskip}
   \end{tabular}
\end{table}

\section*{Analytical Gradient}


Taking the partial derivative of the cost function with respect to the parameter vector lead to:
\begin{align}
\frac{ \partial J( \col{p} ) }{ \partial \col{p} } 
&= \sum R_i \frac{ \partial c_i }{ \partial \col{p} } 
- 2 (\col{p}-\hat{\col{p}})^T \; Q
\end{align}
where with can expand the partial derivative of each measurement cost $c_i$:
\begin{align}
\frac{ \partial c_i }{ \partial \col{p} } = 
\frac{ \partial c_i }{ \partial d_i } \;
\frac{ \partial d_i }{ \partial \col{e}_{ij^*} } \;
\frac{ \partial \col{e}_{ij^*} }{ \partial \col{p} }
\end{align}
where $j^*$ refer to the model point and $i$ to the LiDAR points, and:
\begin{align}
\frac{ \partial c_i }{ \partial d_i } &= 
\frac{
2 d_i 
}{
\ln{10} \left[ 1 + \left( d_i \right)^{2} \right]
}
\\
\frac{ \partial d_i }{ \partial \col{e}_{ij^*} } &= 
\frac{ \col{e}_{ij^*}^T }{ d_i }
\\
\frac{ \partial \col{e}_{ij^*} }{ \partial \col{p} } &= 
\begin{bmatrix}
\frac{ \partial \col{e}_{ij^*} }{ \partial x_o }\;
\frac{ \partial \col{e}_{ij^*} }{ \partial y_o }\;
\frac{ \partial \col{e}_{ij^*} }{ \partial z_o }\;
\frac{ \partial \col{e}_{ij^*} }{ \partial \psi }\;
\frac{ \partial \col{e}_{ij^*} }{ \partial a }\;
\frac{ \partial \col{e}_{ij^*} }{ \partial \Delta_1 } \hdots \frac{ \col{e}_{ij^*} }{ \partial \Delta_l }
\end{bmatrix}
\end{align}
Here, the error is defined by eq. \ref{eq:errorx} with the index $k$ fixed to the index of the closest line to avoid discontinuities in the gradient:
\begin{align}
\col{e}_{ij^*} = \col{e}_{ik^*}^c 
\quad\quad\quad
k^* &= 
\operatornamewithlimits{argmin}\limits_{k} 
\| \col{e}_{ik}^c  \|
\end{align}

Taking the partial derivative of this gradient expression with respect to all parameters lead to:
\begin{align}
\frac{ \partial \col{e}_{ij^*} }{ \partial x_o } &= 
\begin{bmatrix}
0 \\
s\psi \\
-\frac{ \partial z_j }{ \partial x_o} \\
\end{bmatrix}
\quad
\frac{\partial \col{e}_{ij^*} }{ \partial y_o } = 
\begin{bmatrix}
0 \\
-c\psi \\
-\frac{ \partial z_j }{ \partial y_o} \\
\end{bmatrix}
\quad
\frac{\partial \col{e}_{ij^*} }{ \partial z_o } = 
\begin{bmatrix}
0 \\
0 \\
-1 \\
\end{bmatrix}
\\
\frac{ \partial \col{e}_{ij^*} }{ \partial \psi}  &= 
\begin{bmatrix}
0 \\
-(x_i - x_o) c\psi - ( y_i - y_o ) s\psi \\
-\frac{ \partial z_j }{ \partial \psi}
\end{bmatrix}
\\
\frac{ \partial \col{e}_{ij^*} }{ \partial a } &= 
\begin{bmatrix}
0 \\
0  \\
-\frac{ \partial z_j }{ \partial a} \\
\end{bmatrix}
\\
\frac{ \partial \col{e}_{ij^*} }{ \partial h_l } &= 
\begin{bmatrix}
0 &  0 & 0  \\
0 & -1 & 0  \\
-\frac{ \partial z_j }{ \partial x_{k^*} } &  0 & -1 \\
\end{bmatrix}
\begin{bmatrix}
\\
\frac{ \partial \col{r}_{k^*} }{ \partial \Delta_l }
\\ \\
\end{bmatrix}
\quad l \in \{1, \hdots, l_n\}
\end{align}
where 
\begin{align}
\frac{ \partial z_j }{ \partial x_o } &= \frac{ \partial z_j }{ \partial x_j } \frac{ \partial x_j }{ \partial x_o }
\\
\frac{ \partial z_j }{ \partial y_o } &= \frac{ \partial z_j }{ \partial x_j } \frac{ \partial x_j }{ \partial y_o }
\\
\frac{ \partial z_j }{ \partial \psi } &= \frac{ \partial z_j }{ \partial x_j } \frac{ \partial x_j }{ \partial \psi }
\\
\frac{ \partial z_j }{ \partial x_{k^*} } &= \frac{ \partial z_j }{ \partial x_j } \frac{ \partial x_j }{ \partial x_{k^*} }
\end{align}
with
\begin{align}
\frac{ \partial z_j }{ \partial a } &= \cosh{\frac{ x_j }{ a }} - \frac{ x_j }{ a } \sinh{\frac{ x_j }{ a }} - 1 \\
\frac{ \partial z_j }{ \partial x_j } &= \sinh{ \frac{ x_j }{ a }}
\end{align}
and
\begin{align}
\frac{ \partial x_j }{ \partial x_o } &= -c\psi \\
\frac{ \partial x_j }{ \partial y_o } &= -s\psi \\
\frac{ \partial x_j }{ \partial \psi } &= -s\psi ( x_i - x_o ) + c\psi (y_i - y_o ) \\
\frac{ \partial x_j }{ \partial x_k } &= -1
\end{align}
Note that $\frac{ \partial \col{r}_{k^*} }{ \partial \Delta_l }$ is a constant vector provided by the model, corresponding to the $k^*$ column of the partial derivative of the model matrix with respect to $\Delta_l$, see eq. \eqref{eq:matrixmodel}.

\end{document}